\documentclass[lettersize,journal]{IEEEtran}
\usepackage{amsmath,amsfonts}
\usepackage{algorithm}
\usepackage{algorithmic}
\usepackage{array}
\usepackage{booktabs} 
\usepackage[caption=false,font=normalsize,labelfont=sf,textfont=sf]{subfig}
\usepackage{color}
\usepackage{colortbl}
\usepackage{multirow}
\usepackage[table]{xcolor}
\usepackage{textcomp}
\usepackage{stfloats}
\usepackage{url}
\usepackage{verbatim}
\usepackage{graphicx}
\usepackage{cite}
\usepackage{hyperref}
\hypersetup{colorlinks=true, citecolor=green}

\definecolor{lightgray}{RGB}{200,200,200}

\begin{document}

\title{Hierarchical Learning-Enhanced MPC for Safe Crowd Navigation with Heterogeneous Constraints}

\author{Liu Huajian$^{\dagger}$, Feng Yixuan$^{\dagger}$, Dong Wei~\IEEEmembership{Senior Member,~IEEE,} Fan Kunpeng, Wang Chao and Gao Yongzhuo$^{*}$~\IEEEmembership{Member,~IEEE}
\thanks{$\dagger$: Equal contribution.}%
\thanks{All the authors are with State Key Laboratory of Robotic and System, Harbin Institute of Technology, China ($*$: Corresponding author).}
\thanks{This paper was produced by the IEEE Publication Technology Group. They are in Piscataway, NJ.}
\thanks{Manuscript received April 19, 2021; revised August 16, 2021.}}

\markboth{Journal of \LaTeX\ Class Files,~Vol.~14, No.~8, August~2021}%
{Shell \MakeLowercase{\textit{et al.}}: A Sample Article Using IEEEtran.cls for IEEE Journals}


\maketitle

\begin{abstract}
In this paper, we propose a novel hierarchical framework for robot navigation in dynamic environments with heterogeneous constraints. Our approach leverages a graph neural network trained via reinforcement learning (RL) to efficiently estimate the robot’s cost-to-go, formulated as local goal recommendations. A spatio-temporal path-searching module, which accounts for kinematic constraints, is then employed to generate a reference trajectory to facilitate solving the non-convex optimization problem used for explicit constraint enforcement. More importantly, we introduce an incremental action-masking mechanism and a privileged learning strategy, enabling end-to-end training of the proposed planner. Both simulation and real-world experiments demonstrate that the proposed method effectively addresses local planning in complex dynamic environments, achieving state-of-the-art (SOTA) performance. Compared with existing learning-optimization hybrid methods, our approach eliminates the dependency on high-fidelity simulation environments, offering significant advantages in computational efficiency and training scalability. The code will be released as open-source upon acceptance of the paper.
\end{abstract}


\section{Introduction}
\IEEEPARstart{D}{ynamic} scenarios with heterogeneous constraints represent typical operational environments for autonomous mobile robots, including but not limited to airport logistics, hotel delivery, and autonomous driving. Due to the presence of dynamic obstacles, it is challenging for robots to plan an optimal global trajectory even when kinematic constraints are disregarded, as existing planning methods suffer from significant inaccuracies in estimating the cost-to-go, which significantly impacts the autonomous navigation performance of robots in such scenarios.

Most early works did not explicitly distinguish dynamic obstacles, instead relying on high-frequency replanning for obstacle avoidance. However, these approximately greedy local strategies significantly constrain the optimality of the final global trajectory and are highly susceptible to the Freezing Robot Problem due to non-reciprocal interactions with other agents. To more accurately estimate the cost-to-go, researchers have adopted various methods to explicitly predict the future states of dynamic agents, then utilizing approaches such as spatio-temporal joint search to compute the optimal route. Nevertheless, the computational complexity of these methods is excessively high, rendering them applicable only in road scenarios with a very limited number of agents.

Learning-based methods are capable of implicitly estimating the future cost-to-go, thereby achieving enhanced navigation performance in dynamic scenarios. However, these methods lack explicit constraint guarantees, making it challenging to ensure safety when encountering corner cases. Recent studies have attempted to combine learning with optimization to balance performance with interpretable safety. Still, they face significant challenges such as high training costs and sim-to-real transfer issues. Some methods rely on sparse sensors like single-line LiDAR to reduce training costs, but these are generally insufficient for handling complex dynamic scenarios.

To address the aforementioned key challenges in autonomous navigation, this paper proposes a novel hybrid framework based on agent-level features. A graph neural network (GNN)-based frontend is designed to aggregate heterogeneous environmental information and generate step-wise local goals to guide robot navigation, thereby enabling an indirect estimation of the cost-to-go. Meanwhile, explicit dynamic obstacle trajectory prediction and spatio-temporal joint search are incorporated to refine the neural network output, which then serves as a well-initialized input for the model predictive control (MPC)-based backend. As a result, the proposed method achieves superior autonomous navigation performance in dynamic environments.

More importantly, we propose an incremental action masking mechanism and a privileged learning approach to address the issue of infeasible solutions in the backend, which arises when exploring unreachable candidate local goals during early training stages.  This issue results in a large number of timeout episodes, exacerbating the problem of sparse rewards. Leveraging these enhancements, we enable low-cost and highly efficient end-to-end reinforcement learning-based training of the entire planner in a low-fidelity simulation environment. The contributions of this paper can be summarized as follows:

1) We proposed a novel robot navigation framework integrating DRL with MPC for dynamic heterogeneous constraint environments, which effectively combines the implicit reasoning capability of learning-based methods for dynamic scene variations with the explicit constraint guarantees of optimization-based approaches.

2) We design a path search algorithm that incorporates explicit trajectory prediction and spatio-temporal joint graph search, enabling effective dimensionality lifting of the local goals generated by the neural network frontend to yield high-quality initializations for backend optimization.

3) We develop a privileged learning method that utilizes conditional explicit multi-stage simulation of dynamic agents during training, enabling trajectory-level reward evaluation. This method mitigates the modal inconsistency in replanning caused by the lack of temporal regularization.

4) We proposed an incremental action masking mechanism to address the infeasibility issues arising from the exploration of unreachable candidate goals in the early stages. This effectively mitigates the sparse reward problem and enables end-to-end training of the entire planner.
\section{Related Works}
\subsection{Local Planning in Dynamic Environment}
Limited by the perception capabilities, early planning approaches, represented by A*, RRT, DWA \cite{fox1997dynamic} and their variants, primarily focused on static environments. These methods take occupancy grid maps as inputs and can be extended to low-dynamic environments to a limited extent through high-frequency replanning. However, they remain ineffective in handling higher-speed obstacles or environments with more complex interactions. With the improvement in perception technology, various approaches \cite{schmid2023dynablox, xu2025lv, mao2024ram} have been developed to detect and extract dynamic objects within a scene. This enables modeling dynamic obstacles at the agent level rather than the occupancy level. Based on this, model-based methods such as social force models \cite{helbing1995social} , velocity obstacles (VO) \cite{van2011reciprocal}, and the intelligent driver model \cite{treiber2000congested} have been applied to robotic systems. These methods simplify the complex interaction mechanisms between dynamic agents and adopt a single-step reactive policy to facilitate navigation in dense crowds. The key challenge for these models is that they heavily rely on hand-crafted functions and cannot generalize well to various scenarios. In addition, due to their locally greedy nature, the resulting global trajectories often deviate significantly from the optimal path, potentially leading to the Freezing Robot Problem when assumptions such as the reciprocity rule no longer hold \cite{sathyamoorthy2020frozone}. 

Optimization-based methods have become the mainstream in the industry due to their ability to explicitly enforce nonlinear constraints. encompassing not only obstacle avoidance but also the robot’s own kinematics. LMPCC \cite{brito2019model} is a representative hard-constrained method for dynamic environments, which estimates and predicts pedestrian states using a Kalman filter and optimizes the robot’s local trajectory within a model predictive contouring control framework. Gradient-based methods \cite{wang2021autonomous, xu2022dpmpc, lu2024fapp} incorporate penalties for dynamic obstacles into polynomial trajectory optimization, improving computational efficiency at the cost of relaxing constraint guarantees. DPMPC-Planner \cite{xu2022dpmpc} adopts chance-constrained MPC to handle measurement and obstacle uncertainties. \cite{jian2023dynamic} introduces dynamic control barrier functions to enable safety-critical dynamic obstacle avoidance. By leveraging the differential flatness property of car-like robots, \cite{han2023efficient} achieves obstacle avoidance with full-dimensional polygonal constraints, generating less conservative trajectories with safety guarantees. Compared to model-based methods, optimization-based approaches can achieve optimality within a local time domain rather than a single step. However, they remain constrained by computational power and perception range, preventing direct coverage of long-term navigation goals. Consequently, these methods rely on high-level planners to provide reference paths, which, due to the computational complexity, cannot fully incorporate the costs of dynamic obstacles, severely limiting the optimality of global trajectories.
\subsection{Learning-based Crowd Navigation}
Benefits from the implicit modeling capabilities of data-driven approaches, deep neural networks have been utilized to model the complex interactions among agents. Building on this foundation, DRL has emerged as a mainstream solution for crowd navigation problems and can be categorized into sensor-based and agent-based approaches based on the inputs. The former \cite{fan2020distributed, huang2021towards, song2023learning} directly processes raw sensor data, such as images from cameras, while the latter relies on an external perception module to extract features such as agent positions and velocities. Given the relevance to our work, we focus exclusively on agent-based methods which offer significantly lower training costs and robust sim-to-real transfer adaptability, compared to sensor-based counterparts.

Early agent-based methods \cite{chen2017socially, everett2018motion} employ Long Short-Term Memory to aggregate a variable number of agents, addressing the inherent randomness in dynamic scenarios. LM-SARL \cite{chen2019crowd} utilizes a coarse-grained local map to encode interactions between pedestrians. RGL \cite{chen2020relational} adopts a pairwise similarity function to infer correlations between agent features and leverages graph convolutional networks to extract scene features from the resulting graph. \cite{liu2021decentralized, liu2023intention} employ a spatio-temporal interaction graph to model interactions between the robot and surrounding agents, introducing explicit temporal information to reduce the reliance on external velocity estimation of the dynamic agents. However, all these methods adhere to an unrealistic assumption: an infinitely large workspace with homogeneous agents only. To address structured constraints, DRL-VO \cite{xie2023drl} maps VO-based agent features onto a feature map, which is combined with multi-frame LiDAR maps to construct an image-based unified input representation. HEIGHT \cite{liu2024height} generalizes \cite{liu2023intention} to highly constrained environments by incorporating independent LiDAR features. 

To provide a consistent representation of agents and structured obstacles without relying on raw LiDAR measurements, \cite{bachiller2022graph} models dynamic scenes using heterogeneous graphs, while \cite{han2022reinforcement} extends VO to unify the node features of heterogeneous entities. Building upon this, \cite{zhou2022navigating} employs graph attention networks to extract features from heterogeneous graphs, constructing end-to-end policy networks for autonomous navigation in dynamic environments. Compared to LiDAR-fusion methods, heterogeneous-graph-based approaches do not require high-fidelity simulation environments, offering a significant advantage in terms of training costs. Moreover, with the support of semantic-level perception methods such as BEVFormer \cite{li2024bevformer}, these planners have become practically viable. However, despite their theoretical advantages, these end-to-end methods lack explicit constraint guarantees, and their inherent opacity makes them less reliable when handling corner cases, compromising robustness and safety.
\subsection{Combining Optimization with Learning}
The integration of optimization and learning has been extensively explored in autonomous robotics research, as comprehensively reviewed in \cite{hewing2020learning}. Intuitively, many existing works leverage the implicit modeling capabilities of neural networks to address inherent limitations in optimization-based methods, such as inaccurate modeling, sensitivity to initial conditions, and the reliance on manually tuned objective functions. Conversely, optimization techniques have also been employed to accelerate neural network training and to serve as safety filters for learning-based methods \cite{wabersich2021predictive}. Within the scope of this paper, we specifically focus on methods where the optimization-based planner serves as the final output carrier.

GO-MPC \cite{brito2021go} proposes a hybrid method based on local goal recommendation to help the robot in making progress towards its goal and accounts for the expected interaction with other agents. Similarly, \cite{lim2024learning} determines the terminal state and confidence of the trajectory using DRL for road scenarios and subsequently plans a QP trajectory based on them. By placing a differentiable MPC \cite{amos2018differentiable} as the last layer of the actor network, AC-MPC \cite{romero2024actor} effectively manages both short-term decisions through the MPC-based controller and long-term predictions via the critic network. Notably, iPlanner \cite{yang2023iplanner, roth2024viplanner} proposes imperative learning, a novel approach to obtain the gradients for network training while simultaneously optimizing output trajectories. This method leverages a differentiable cost map to provide implicit supervision, eliminating the need for demonstrations or labeled trajectories. NeuPAN \cite{han2025neupan} employs an interpretable deep unfolding network to directly obtain latent distance features from raw LiDAR clouds, thereby replacing obstacle avoidance constraints with a computationally efficient regularization term in the loss function. However, these methods do not effectively handle both complex multi-agent interactions and environmental structural constraints simultaneously. Moreover, most existing methods do not explicitly account for the robot’s kinematic constraints.  

The works most closely related to this paper are GO-MPC \cite{brito2021go} and AC-MPC \cite{romero2024actor}. Specifically, both the proposed method and GO-MPC utilize local goals as an intermediary between the front-end network and the back-end optimization to estimate the cost-to-go. However, GO-MPC is only applicable in environments without structural constraints and does not account for the robot’s kinematic constraints. Conversely, AC-MPC focuses solely on environmental constraints and external disturbances but lacks the ability to handle dynamic obstacles. Combining the advantages of both, the proposed method integrates structural constraints, dynamic obstacle avoidance, and the robot’s  kinematic constraints while supporting end-to-end training in low-fidelity simulation environments.

\section{Methodology}
\subsection{Preliminaries and Problem Formulation}
\subsubsection{Problem formulation}
\label{sec:problem}
Consider an episodic environment in a 2D manifold $\mathcal{W} \in \mathbb{R}^2$, where an autonomous mobile robot must navigate from an initial state $\mathbf{x}_0$ to a goal $\mathbf{g}$. During this process, the robot needs to interact with pedestrians and heterogeneous static obstacles. Motion planning of the robot can be formulated as an optimal control problem:
\begin{equation}
    \begin{split}
    J^*(\mathbf{x}(0)) = \min_{\substack{ \\ \{\mathbf{u}(\tau)\}_0^{t_f}}} \biggl\{ & c_f(\mathbf{x}(t_f), t_f) \\
    & + \int_{0}^{t_f} c(\mathbf{x}(\tau), \mathbf{u}(\tau), \tau) \, d\tau \biggr\}
\end{split},
\label{eq:ocp}
\end{equation}
where $\mathbf{x}(0)$ is the initial state and $\mathbf{x}(t_f)$ is the terminal state. The solution of (\ref{eq:ocp}) can be described in the form of an optimal trajectory $\mathcal{T}^* := \mathbf{x}^*(\tau): [0, t_f]$. However, since the cost function $c(\mathbf{x}(t), \mathbf{u}(t), t)$ is not explicitly accessible in dynamic environments, global optimal policy $\pi^*_G(\mathbf{x}(t), t)$ for robot navigation cannot be easily obtained by solving (\ref{eq:ocp}). Fortunately, based on the principle of optimality, for any point $\mathbf{x}^*(t^{\prime})$ on the optimal trajectory $\mathcal{T}^*$ within the neighborhood of the robot, the truncated trajectory $\mathbf{x}^*(\tau) : [t^{\prime},t_f]$ is also a minimum cost trajectory whose cost-to-go can be denoted as $J^*(\mathbf{x}(t^{\prime}))$. Benefiting from this, the primal motion planning problem can be decomposed into the following two subproblems:

\textbf{Problem 1}: Find the local waypoint $\mathbf{p}_t^{\text{ref}}$ with the minimal cost-to-go $J^*(\mathbf{p}_t^{\text{ref}})$ in the robot's perceptual range. 

\textbf{Problem 2}: Formulate and solve an optimization problem with $\mathbf{p}^{\text{ref}}_t$ as the terminal state, considering obstacle avoidance and kinematic constraints, to obtain the locally optimal control policy $\pi^*_L(\mathbf{S}_t, \mathbf{p}^{\text{ref}}_t)$, where $\mathbf{S}_t$ represents the joint observable state at time $t$.

Since problem 1 is still directly intractable, we utilize a learnable policy neural network to estimate $J^*(\mathbf{p}^{\text{ref}}_t)$ based on $\mathbf{S}_t$, then Problem 1 can be reformulated as: to learn a policy $\pi_{\theta}$ to recommend local goal $\mathbf{p}^{\text{ref}}_t$ for the robot that maximizes the expected return, while ensuring collision-free and dynamic-feasible. We formally described the discrete form of the reformulated problem as follows:
\begin{equation}
    \begin{aligned}
	\pi_{\theta}^* = & \arg\max_{\theta} \mathbb{E}\left[\sum_{t=0}^T \gamma^t R\left(\mathbf{S}_t, \pi^*_L(\mathbf{S}_t, \pi_{\theta}(\mathbf{S}_t))\right)\right] \\
	\text{s.t.} \quad & \mathbf{x}_{t+1} = f_d(\mathbf{x}_t, \mathbf{u}_t) \\
	& \|\mathbf{p}_T - \mathbf{g}\| \le \epsilon \\
	& \mathcal{O}_t(\mathbf{x}_t) \cap \mathcal{O}_i^t = \emptyset \\
	& \mathbf{u}_t \in \mathcal{U}, \; \mathbf{S}_t \in \mathcal{S}, \; \mathbf{x}_t \in \mathcal{X}, \\ 
	&\forall t \in [0, T], \forall i \in \{1, \ldots, n\},
    \end{aligned}
\end{equation}
where $\gamma \in (0, 1]$ is a discount factor and also serves as a tunable hyper-parameter, $\mathcal{O}_t$ and $\mathcal{O}_t^i$ represent the time-varying spatial occupancy of the robot and obstacles in $\mathcal{W}$, respectively.
\subsubsection{Robot kinematics}
Unlike simplified RL formulations that assume idealized omnidirectional kinematic models, practical planners must account for the inherent kinematic constraints of physical hardware and discretization errors introduced during implementation. To address this gap, we adopt the industry-standard differential-drive robot as a representative platform and model its dynamics through a differential equation $\dot{\mathbf{x}} = f(\mathbf{x}, \mathbf{u})$, given by:
\begin{equation}
    \begin{aligned}
	& \dot{x} = \frac{v_r + v_l}{2} \cos \psi \quad \dot{v}_r = a_r \\
	& \dot{y} = \frac{v_r + v_l}{2} \sin \psi \quad  \dot{v}_l = a_l \\
	& \dot{\psi} = \frac{v_r - v_l}{2 \rho} 
\end{aligned},
\label{eq:dyna}
\end{equation}
where $x$ and $y$ are the robot's position coordinates and $\psi$ is the heading angle in a global frame of $\mathcal{W}$. $v_l$ and $v_r$ represent the velocities of the wheels on the left and right side of the robot respectively, while $a_l$ and $a_r$ correspondingly denote the accelerations. Furthermore, we denote the hardware constraints on the velocity and acceleration of each wheel as $[-v_{\text{max}}, v_{\text{max}}]$ and $[-a_{\text{max}}, a_{\text{max}}]$, respectively. Based on these definitions and analysis, the control vector can be denoted as $\mathbf{u} = (a_l, a_r)$, while the set of admissible control states $\mathcal{U}$ is defined as $([-a_{\text{max}}, a_{\text{max}}], [-a_{\text{max}}, a_{\text{max}}])$.
\subsection{Scenario to Markov Decision Process}
Navigating a robot toward the goal $\mathbf{g}$ by generating optimal local goals $\mathbf{p}_t^{\text{ref}}$ in dynamic environments can be regarded as a sequential decision-making problem. We model this problem as a Markov Decision Process (MDP) $\mathcal{M}$ and describe it using a tuple $\langle \mathcal{S}, \mathcal{A}, P, R, \gamma \rangle$. Here, $\mathcal{S}$ and $\mathcal{A}$ represent the observable state space and action space, respectively. $P: \mathcal{S} \times \mathcal{A} \rightarrow \mathcal{S}$ denotes the unknown state transition function. $\gamma$ is the discount factor for balancing short-term and long-term rewards.

Based on our previous work, SAGE \cite{huajian2024sample}, we classify the instances in the scene into four classes: ego-robot, pedestrian, static circular obstacle, and line obstacle (large polygon obstacles can be decomposed into a set of line obstacles). Utilizing the velocity obstacle (VO) vector \cite{han2022reinforcement}, we define the feature vectors for each class as follows:
\begin{equation}
    \mathbf{h} = \begin{cases}
        [d_g, \mathbf{v}, v_m, \psi] \in \mathbb{R}^5, \; &\text{(robot)}  \\
        [\mathbf{v}, \mathbf{v}_l, \mathbf{v}_r, \mathbf{p}, \tilde{\rho}, \mu, \zeta] \in \mathbb{R}^{11}, \;  &\text{(pedestrian)} \\
        [\mathbf{v}_l, \mathbf{v}_r, \mathbf{p}, \tilde{\rho}, \mu, \zeta] \in \mathbb{R}^9, \; &\text{(circular obstacle)} \\
        [\mathbf{v}_l, \mathbf{v}_r, \mathbf{p}_s, \mathbf{p}_e, \rho, \mu, \zeta] \in \mathbb{R}^{10}, \; &\text{(line obstacle)}
    \end{cases},
    \label{eq:feature}
\end{equation}
where $d_g$ is the Euclidean distance between the robot's position and its goal, the unsubscripted $\mathbf{v}$ denotes the velocity vector of a movable entity, $v_m$ is the robot’s maximum linear velocity, $\mu$ is the minimum distance between the entity and the robot, $\tilde{\rho}$ is the radius of the pedestrian or static circular obstacle enlarged by the robot's radius $\rho$, and ($\mathbf{p}_s$, $\mathbf{p}_e$) represent the start and end positions of the line obstacle, respectively.

\begin{figure}[htbp]
    \centering
    \includegraphics[width=0.48\textwidth]{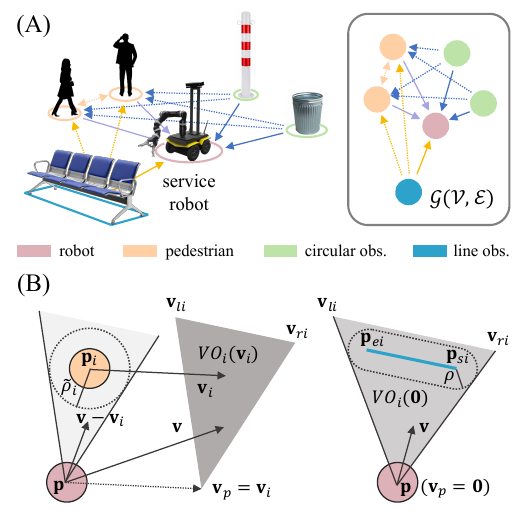}
    \caption{(a) Transformation from a crowd scenario to a heterogeneous graph; (b) Illustration of the velocity obstacle for pedestrians and static line obstacles, symbols without subscripts represent the state variables of the ego-robot (Reprinted from our previous work \cite{huajian2024sample}, with permission from IEEE.).\label{fig:crowd_graph}} 
\end{figure}

In addition to the aforementioned conventional state variables, vector $[\mathbf{v}_i, \mathbf{v}_{li}, \mathbf{v}_{ri}] \in \mathbb{R}^6$ defines a VO cone between the robot and the $i$-th obstacle, as shown in Fig. \ref{fig:crowd_graph}, while $\zeta = 1/(\xi + 0.5)$ is an artificial state variable quantifying collision risk, where $\xi$ denotes the expected collision time computed based on the VO cone. Using the observable features described in (\ref{eq:feature}), we model the scene as a heterogeneous graph, which serves as the state input $\mathbf{S} \in \mathcal{S}$ in $\mathcal{M}$. Each node in the graph corresponds to an entity in the scene, while the edges represent implicit interactions between connected nodes.

As described in Section \ref{sec:problem}, we define the local goals as the action $\mathbf{a} \in \mathcal{A}$ in $\mathcal{M}$. The reward function $R(\mathbf{S}, \mathbf{a}): \mathcal{S} \times \mathcal{A} \rightarrow \mathbb{R}$ is designed roughly following \cite{huajian2024sample}. To ensure safety and task completion, two sparse reward terms, $r_c$ and $r_g$, are defined. The former penalizes any collision involving the ego-robot, while the latter is awarded when the robot successfully reaches the goal. To enhance navigation efficiency, a step-dependent reward term, $r_t$, is introduced to penalize excessive navigation time. At any given training step, the robot receives exactly one of these rewards, denoted as $R_s \in \{r_c, r_g, r_t\}$. Drawing inspiration from recent works \cite{han2022reinforcement}, $r_{\psi}$ is incorporated to guide the robot's orientation towards the goal under nonholonomic constraints, and $r_v$ is used to account for potential collisions under the current configuration. Furthermore, a dense social-aware reward, $r_s$, discourages the robot from approaching pedestrians too closely, improving social compliance. Then, $R(\mathbf{S}, \mathbf{a})$ is defined as the sum of the following components:
\begin{equation}
    \begin{aligned}
        &  R_s = \begin{cases} 
            \kappa_1, & \text{if \;} d_g < \epsilon \\
            \kappa_2, & \text{if \;collision happens} \\
            \kappa_3, & \text{otherwise}
        \end{cases} \\
        & r_{\psi} = \kappa_4(\cos(\psi_g) - 1) / (d_g + d_{\psi}) \\
        & r_v = \sum_{\xi_i < \xi_c} (\kappa_5 + \kappa_6 \zeta_i) \\
        & r_s = \kappa_7 \sum_{\text{type(i) is ped.}} \min\{0, \mu_i - \mu_s\}
    \end{aligned},
    \label{eq:reward}
\end{equation}
where $\epsilon$ is a small convergence tolerance for determining if the goal has been reached. $\mu_s$ is a user-defined safety margin, $\psi_g$ is the angle between the robot's orientation and the vector pointing from its current position to the goal, and $d_{\psi}$ is a hyper-parameter adopted for smoothing and bounding $r_{\psi}$. $\xi_c$ is a specific threshold designed to distinguish whether a collision risk exists. The hyper-parameters $\kappa$ in the reward function represent tunable weight coefficients.
\subsection{Neural Network Architecture}
\label{sec:network}
\begin{figure}[htbp]
    \centering
    \includegraphics[width=0.48\textwidth]{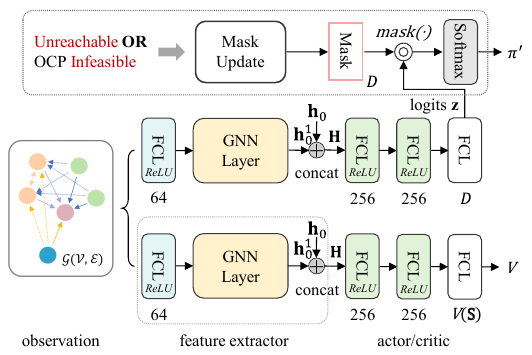}
    \caption{An illustration of the architecture of the proposed GNN-based policy network with invalid action masking mechanism.\label{fig:net_arch}} 
\end{figure}

Building upon the classical actor-critic architecture, we design the policy and value networks using graph neural network layers as the feature extractors, as illustrated in Fig. \ref{fig:net_arch}. Based on empirical results, we do not employ a shared feature extraction layer as similar works do. The output of each GNN is concatenated with the robot's node feature to form the integrated fixed-length observation $\mathbf{H}$. Both the policy network $\pi_{\theta}$ and the value network $V$ consist of two hidden fully connected layers (FCLs) with 256 rectified linear units.

The key difference from related works is that we adopt a discrete action space instead of a continuous counterpart. Specifically, we discretize the robot’s neighborhood into $D$ candidate points and employ a static bijective function to map these candidates to the one-dimensional categorical output of the network, $a \in \mathcal{A} := [1, D]$. Based on this formulation, we use a fully connected layer with $D$ units as the final output layer of the policy network, producing unnormalized scores (logits) $z \in \mathbb{R}^D$. Finally, a softmax function is applied to transform the logits into an action probability distribution:
\begin{equation}
    \pi_{\theta}(a | \mathbf{S}) = \frac{e^{z_a}}{\sum_{d = 1}^D e^{z_d}},
    \label{eq:softmax}
\end{equation}
where $\pi_\theta(a | \mathbf{S})$ represents the probability of the policy network selecting the local goal corresponding to action $a$ given state $\mathbf{S}$, where $\theta$ denotes the learnable parameters of the network.

The motivation behind this architecture is that the chosen local goal $\mathbf{p}_{\text{ref}}$ may render the optimization problem $\pi_L(\mathbf{S}, \mathbf{p}_{\text{ref}})$ infeasible, which could destabilize training within the RL framework. Inspired by \cite{vinyals2017starcraft}, we design a simple yet effective incremental invalid action mask to prevent repeatedly sampling invalid actions in relatively large discrete action spaces. Specifically, we define the following two types of local goals as invalid actions: 1) spatially unreachable (e.g. located inside a static obstacle); 2) spatially reachable but rendering $\pi_L(\mathbf{S}, \mathbf{p}_{\text{ref}})$ infeasible. We mask out invalid candidate local goals by replacing the corresponding logits $z$ with a large negative number (e.g. $-1 \times 10^8$). Then the updated policy, denoted as $\pi_{\theta}^{\prime}$, can be obtained by recalculating (\ref{eq:softmax}). The resulting probability of an invalid candidate point, denoted as $\epsilon_m$, should be a negligibly small value close to zero. 

The masking process, denoted as $mask(\cdot) : \mathbb{R} \rightarrow \mathbb{R}$, can be viewed as a state-dependent function differentiable to its parameters $\theta$. Consequently, the gradient produced by $mask(\cdot)$ is a valid policy gradient of $\pi_{\theta}^{\prime}$. In addition, $mask(\cdot)$ ensures that the gradient corresponding to the logits of invalid candidate points is set to zero, thereby introducing a form of supervised learning. Notably, the feasibility of $\pi_L$ cannot be determined through predefined conditions. Therefore, we adopt an incremental approach, where infeasible candidate points are progressively added to the set of invalid candidates $\tilde{\mathcal{P}}$, and $mask(\cdot)$ is updated accordingly to recalculate $\pi_{\theta}^{\prime}$. This process can be parallelized with negligible computational overhead to enhance inference efficiency.
\subsection{Spatio-Temporal Path Search}
\label{sec:search}
The robot trajectory optimization problem $\pi_L$ with obstacle avoidance constraints is typically non-convex, and its efficient solution relies on the proper selection of an initial trajectory $\mathcal{T}_{\text{init}}$. To address this, we propose a spatio-temporal (S-T) path searching module based on explicit predictions of dynamic obstacle trajectories, denoted as $\mathcal{P}_{\text{ref}} = G(\mathbf{p}_{\text{ref}})$, to increase the dimension of the output of the high-level RL policy $\pi_{\theta}^{\prime}(\mathbf{S})$. This design effectively reduces the action space $\mathcal{A}$ of the RL policy, significantly lowering the convergence difficulty and training cost compared to approaches that directly output the entire reference trajectory \cite{yu2024pathrl}. Meanwhile, the introduction of local goals substantially reduces the search space of $G(\cdot)$, ensuring feasibility for real-time online computation.

As a direct extension of traditional graph search-based methods, the 2D spatial plane $\mathcal{W}$ is augmented with an orthogonal time axis $t \in \mathbb{R}_{\ge 0}$, allowing the time-varying spatial occupation of dynamic obstacles to be accurately represented. Specifically, based on our previous work on multi-agent trajectory prediction \cite{liu2024pmm}, we can predict the future trajectory $\hat{\mathbf{Y}}$ of an agent using its historical path. Then the Minkowski sum of $\hat{\mathbf{Y}}$ and the agent's spatial geometric shape can be directly projected to the occupancy grid in the S-T map. Due to the positivity of $\dot{t}$, the graph constructed from this occupancy grid will be a directed acyclic graph (DAG). For robots with nonholonomic constraints as described in (\ref{eq:dyna}), the transition cost between grids is directly influenced by the orientation angle $\psi$. Therefore, we use $\mathbf{h}_{st} := [x, y, \psi, t, g, h]$ as the node features in the resulting S-T graph $\mathcal{G}_{\text{ST}}$.

In S-T graph-based search, node expansion is often based on motion primitives to account for kinematic constraints. However, for small discretization intervals $\delta t$, the primitives lack sufficient distinguishability within the discrete grid, leading to a large number of redundant candidate edges, which severely degrades search efficiency. Inspired by Pulse Width Modulation, we address this issue by selecting only the maximum velocity or zero velocity and a fixed $\delta t$ when generating primitives. This results in seven motion primitives, including an in-place waiting action, while primitives for other configurations can be approximated through combinations of these seven. We denote the set formed by them as $\mathcal{M}_p$.

By computing the projection of the terminal state of each primitive onto the S-T grid map, we can obtain adjacent nodes and update the node's orientation $\psi$. Building on this, we define the \textbf{cost function} for any node $n_{st}$ in the S-T graph according to the navigation task requirements as: $J(n_{st}) = g(n_{st}) + h(n_{st})$, where $g(n_{st})$ represents the accumulated cost from the starting point $\mathbf{x}(0)$ to the current node, and $h(n_{st})$ is a heuristic function estimating the cost to reach the destination $\mathbf{p}_{\text{ref}}$. They are defined respectively as follows:
\begin{equation}
    \left\{\begin{aligned}
        & g(n_{st}) = \sum_0^t (d(x, y) + g_r(\Delta x, \Delta y, \psi) + g_{\psi}(\Delta \psi)) \\
        & h(n_{st}) = E(x, y) + \kappa_c C(x, y, t) + h_t(t)
    \end{aligned}\right. ,
    \label{eq:search_cost}
\end{equation}
where $d(x, y)$ represents the Euclidean distance change, whose accumulation approximates the historical path length. Both $g_r$ and $g_{\psi}$ are piecewise constant functions to penalize reversing and turning, respectively. $E(x, y)$ denotes the Euclidean distance between $n_{st}$ and $\mathbf{p}_{\text{ref}}$. $C(x, y, t)$ represents the angle between the vector from $n_{st}$ to $\mathbf{p}_{\text{ref}}$ and the vector from $\mathbf{x}(0)$ to $\mathbf{p}_{\text{ref}}$ in the S-T space, with $\kappa_c$ as the corresponding tunable coefficient. This component encourages node expansion along a straight trajectory at a constant velocity. $h(t)$ is a time-based reward function that prioritizes nodes with larger $t$ values, ensuring real-time performance.

We employ the A-Star algorithm to solve the resulting graph search problem and apply velocity pruning to restrict node expansion, which prevents the exploration of regions in S-T space that violate kinematic constraints, improving search efficiency. After obtaining the optimal path $\mathcal{P}^\star$ in $\mathcal{G}_{\text{ST}}$, we select the corresponding key points from $\mathcal{P}^\star$ based on the time interval $\Delta t$ of the optimization problem to form $\mathcal{P}_{\text{ref}}$. The complete spatio-temporal path search algorithm is summarized in Algorithm \ref{alg:search}.
\begin{algorithm}[htbp]
    \caption{\textbf{S}patio-\textbf{T}emporal  joint path search.}\label{alg:search}
    \begin{algorithmic}[1]
        \REQUIRE S-T graph $\mathcal{G}_{\text{ST}}$, $\mathcal{M}_p$, $\mathbf{p_{\text{ref}}}$, $\Delta t$
        \ENSURE Reference path for the trajectory optimization  $\mathcal{P}_{\text{ref}}$
        \STATE $\mathcal{O} \leftarrow \{Node(\mathbf{x}(0), g: 0, h: h(\mathbf{x}(0)), p: None)\}$ \\   
        \COMMENT{Initialize open list (priority queue sorted by $J$)} 
        \STATE $\mathcal{C} \leftarrow \emptyset$    \COMMENT{Initialize set of closed nodes}
        \WHILE{$\lvert \mathcal{O} \lvert > 0$}
            \STATE $n_{st} \leftarrow$ PriorityQueuePop($\mathcal{O}$) \COMMENT{Pop node with lowest J-value}
            \IF{$\|n_{st}.(x, y) - \mathbf{p}_{\text{ref}}\| \le \epsilon$}
                \STATE $\mathcal{P}^\star \leftarrow$ TraceBack($n_{st}$)
                \RETURN KeyPointSelection($\mathcal{P}^\star$, $\Delta t$)
            \ENDIF
            \FORALL{$m \in \mathcal{M}_p$}
                \item[] $\triangleright$ \textit{Iterate over all seven primitives} 
                \STATE $\mathbf{x}_f^\prime \leftarrow$ Transform($m$, $n_{st}$) \COMMENT{Next state $(x, y, \psi, t)$}
                \IF{\textbf{not} isValid($\mathcal{G}_{\text{ST}}, \mathbf{x}_f^\prime$) \textbf{or} isClosed($\mathcal{C}, \mathbf{x}_f^\prime$)}
                    \STATE \textbf{continue}
                \ENDIF
                \STATE $(g^\prime, h^\prime) \leftarrow$ CostFunction($n_{st}$, $\mathbf{x}_f^\prime$) \COMMENT{(\ref{eq:search_cost})}
                \STATE $n \leftarrow$ OpenListQuery($\mathcal{O}, \mathbf{x}_f^\prime$)
                \IF{isNull($n$)}
                    \STATE $n_{st}^\prime \leftarrow Node(\mathbf{x}_f^\prime, g: g^\prime, h: h^\prime, p: n_{st})$
                    \STATE PriorityQueueInsert($\mathcal{O}, n_{st}^\prime$)
                \ELSIF{$g^\prime < n.g$}
                    \STATE UpdateNode($n, \{g:  g^\prime, p: n_{st}\}$) \COMMENT{Update the node based on a more optimal incoming path}
                \ENDIF
            \ENDFOR
            \STATE Insert $n_{st}$ in $\mathcal{C}$
        \ENDWHILE
        \RETURN Infeasible  \COMMENT{Update invalid action mask} 
    \end{algorithmic}
\end{algorithm}
\subsection{Optimization-based Trajectory Refinement}
\label{sec:ocp}
Using the reference path $\mathcal{P}_{\text{ref}}$ obtained from evaluating $G(\mathbf{p}_{\text{ref}})$, we design a trajectory refinement module within the MPC framework to generate a collision-free locally optimal trajectory $\mathcal{T}_{\text{opt}}$ that satisfies the robot's kinematic constraints. Since model training is performed offline, we trade off some computational efficiency for minimal conservatism.

We adopt a hierarchical approach to enforce obstacle avoidance (OA) constraints. Specifically, for dynamic obstacles, we reuse the predicted trajectory $\hat{\mathbf{Y}}$ from Section \ref{sec:search}. The OA constraints for pedestrians and static circular obstacles can then be defined in terms of the Euclidean distance between the robot and the obstacles at each prediction step $k$. For convex polygonal obstacles derived from decomposed structured obstacles, their spatial distance to the robot can be quantified using signed distance \cite{boyd2004convex}. Leveraging the strong duality property, we reformulate the signed distance into a smooth and differentiable hyperplane separation constraint. Combining OA constraints with the robot’s kinematic constraints (\ref{eq:dyna}), we formulate the following non-convex optimization problem for trajectory refinement:
\begin{equation}
    \begin{aligned}
	   J^* =  &\ \arg \min \sum_{k = 0}^{N - 1}J(\mathbf{x}_k, \mathbf{u}_k, \Delta\mathbf{u}_k, \mathbf{p}_k^{\text{ref}}) + J(\mathbf{x}_N, \mathbf{p}^{\text{ref}}_N) \\
	   \text{s.t.} \quad &\mathbf{x}_{k + 1} = f_{\text{RK4}}(\mathbf{x}_k, \mathbf{u}_k),  \\
	   & \|\mathbf{p}_k - \hat{\mathbf{p}}_k^i\| \ge \tilde{\rho}_i,  \\
	   & (\mathbf{A}_m\mathbf{p}_k - \mathbf{b}_m)^T\boldsymbol{\lambda}_k^{(m)} + s_k^{(m)} > \rho, \\
	   & \| \mathbf{A}_m^T \boldsymbol{\lambda}_k^{(m)}\|_{\ast} = 1, \\
	   & s_k^{(m)} \in \mathbb{R}_{+}, \; \boldsymbol{\lambda}_k^{(m)} \succeq_{\mathcal{K}^*} 0, \\
	   & \mathbf{u}_k \in \mathcal{U}, \; \mathbf{x}_k \in \mathcal{X}, \\
	   & \forall k \in \{0, \dots, N\}, \\
	   & \forall i \in \{1, \dots, n\}, \; \forall m \in \{1, \dots, M\}.
    \end{aligned},
    \label{eq:mpc}
\end{equation}
where $f_{\text{RK4}}$ represents the discrete dynamics equations derived using the fourth-order Runge-Kutta method. $\mathbf{u}_k$ denotes the control input of the robot, which consists of linear acceleration $\mathbf{a}_k^{\text{lin}}$ and angular acceleration $\alpha_k$ for a differential drive robot. $\mathbf{p}^{\text{ref}}_k \in \mathcal{P}_{\text{ref}}$ is the reference path point corresponding to the current stage. The tuple ($\mathbf{A}_m$, $\mathbf{b}_m$) denotes the half-plane representation of the $m$-th polygon obstacle $\mathcal{P}_m$, $\boldsymbol{\lambda}^{(m)}$ is the dual variable and $s^{(m)}$ is the slack variable associated with $\mathcal{P}_m$. The operator $\| \cdot \|_{\ast}$ is the dual norm and $\succeq_{\mathcal{K}^*}$ is the generalized inequality defined by a dual cone $\mathcal{K}^*$. Details of the derivation and proof can be found in \cite{boyd2004convex, liu2024optimization}.

\begin{figure}[htbp]
    \centering
    \includegraphics[width=0.48\textwidth]{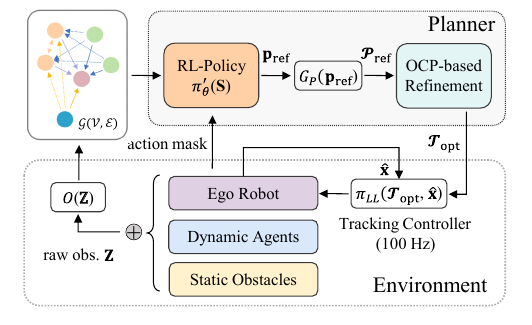}
    \caption{Illustration of the proposed hierarchical planner, where $O(\cdot)$ represents the observation function.\label{fig:sys_arch}} 
\end{figure}
\subsection{Privileged Reinforcement Learning}
The output of the planner transforms into an optimal trajectory $\mathcal{T}_{\text{opt}}$ after refinement. However, from the perspective of the neural network being trained, its output is still the local goal $\mathbf{p}_{\text{ref}}$, while the entire backend is considered part of the environment. This inconsistency creates an adaptability issue for traditional reinforcement learning algorithms for the proposed planner. Specifically, conventional methods optimize the network based on data tuples $\langle \mathbf{S}_{t}, \mathbf{S}_{t+1}, \mathbf{a}_t, r\rangle$, which causes the value network to focus only on the changes between two adjacent observation states $\mathbf{S}_{t}$ and $\mathbf{S}_{t+1}$, and the corresponding reward $r$, while ignoring the subsequent optimal trajectory that has not yet been executed. As a result, the selection of local goals loses temporal consistency, leading to significant inter-frame jitter in the planned trajectory.

To address this issue, we revisit the relationship between the planner and the environment, considering $\mathcal{T}_{\text{opt}}$ as part of the state $\mathbf{S}$, specifically as the current trajectory being executed, denoted as $\tilde{\mathbf{S}}$. With this modification, the state value function $V(\tilde{\mathbf{S}})$ can more accurately evaluate the real situation. Due to the focus of the proposed method on low-fidelity simulation environments, we develop a privileged training approach to jointly train the planner for overall value evaluation of $\mathcal{T}_{\text{opt}}$ instead of $\mathbf{S}$. Specifically, at each rollout step during the training process, we simulate the environment forward by $M$ steps, with $M \leq N$ under the condition of the fixed robot trajectory $\mathcal{T}_{\text{opt}}$. Following (\ref{eq:reward}), we evaluate and record the safety-related rewards $R_s^k = r_c^k + r_s^k + r_v^k, k \in [1, M]$ at each step and simultaneously backtrack the environment to the state at $k = 1$. Finally, the overall reward function for this rollout is defined as follows: 
\begin{equation}
    R(\tilde{\mathbf{S}}) = r_g + r_{\theta} + r_t + \sum \limits_{k = 1}^M \lambda^{k - 1}_p R_s^k
    \label{eq:priv_cost}
\end{equation}
where $\lambda_p \in [0, 1]$ is a tunable hyperparameter representing the discount factor for the influence of future rewards on the current decision. Particularly, collisions occurring at $k > 1$ do not prematurely terminate the current training episode as real collisions do. Most importantly, in contrast to the privileged value network that has access to perfect knowledge of the environmental evolution, the policy network as student can only access the observable state at the current time step, $\mathbf{S}_t$. This design eliminates data leakage and ensures that the policy network generalizes effectively to unseen scenarios.

Since the proposed method is agnostic to the RL algorithm, based on the privileged reward function, it can be trained with any RL algorithm that adapts to continuous state spaces and discrete action spaces with minimal modification. We summarize the entire privileged reinforcement learning algorithm in Algorithm \ref{alg:learning}. Considering that the navigation task is challenging for RL without supervision, we follow the paradigm of curriculum learning and divide the training process into multiple phases to gradually increase the number of agents in the training environments. The specific details will be presented in Section \ref{sec:env}. 

\begin{algorithm}
    \caption{Privileged Reinforcement Learning.}\label{alg:learning}
    \begin{algorithmic}[1]
        \REQUIRE Initial policy parameters $\{ \theta_0, \phi_0 \}$, environment $\mathcal{M}$
        \ENSURE Optimal parameters $\theta^*$ of policy network $\pi_{\theta}$
        \WHILE{$episode < n_{\text{episode}}$}
            \FOR{$k = 0, \ldots, n_{\text{mini-batch}}$}
                \STATE $\mathcal{T}_{\text{opt}} \leftarrow \emptyset$, $\tilde{\mathcal{P}} \leftarrow \emptyset$  \COMMENT{Initialize invalid action set}
                \STATE $\hat{\mathbf{Y}} \leftarrow$ TrajectoryPrediction($\mathbf{Z}$) \COMMENT{$\mathbf{Z}$ denotes the raw vector-based observation}
                \STATE $\mathbf{S} \leftarrow$ SceneToGraph($\mathbf{Z}$)
                \WHILE{$iter$++ $< n_{\text{iter}}$}
                    \STATE $\pi_{\theta}^\prime \leftarrow$ UpdateInvalidActionMask($\pi_{\theta}, \mathbf{Z}, \tilde{\mathcal{P}}$)
                    \STATE $\mathbf{p}_{\text{ref}} \sim \pi_\theta^\prime(\mathbf{S})$   \COMMENT{Sample from the updated probability distribution}
                    \STATE $\mathcal{P}_{\text{ref}} \leftarrow G(\mathbf{p}_{\text{ref}}, \hat{\mathbf{Y}})$ \COMMENT{Via Algorithm \ref{alg:search}}
                    \IF{infeasible}
                        \STATE $\tilde{\mathcal{P}} \leftarrow \tilde{\mathcal{P}} \cup \{\mathbf{p}_{\text{ref}}\}$    \COMMENT{Update invalid action set}
                        \STATE \textbf{continue}
                    \ENDIF
                    \STATE $\mathcal{T}_{\text{opt}} \leftarrow $ Solve (\ref{eq:mpc}) with $\{\mathcal{P}_{\text{ref}}, \mathbf{Z}, \hat{\mathbf{Y}}\}$
                    \IF{\textbf{not} success}
                        \STATE $\tilde{\mathcal{P}} \leftarrow \tilde{\mathcal{P}} \cup \{\mathbf{p}_{\text{ref}}\}$
                        \STATE \textbf{continue}
                    \ENDIF
                    \STATE \textbf{break}   \COMMENT{Local optimal trajectory obtained}
                \ENDWHILE
                \IF{$\lvert \mathcal{T}_{\text{opt}} \lvert < N$}
                    \STATE $\mathbf{u}^\prime \leftarrow [0, 0]$ \COMMENT{No solution found, braking}
                    \STATE $\{\mathbf{S}, \mathbf{S}^\prime, \mathbf{p}_{\text{ref}}, r, done \} \leftarrow$ $\mathcal{M}$.Step($\mathbf{u}^\prime$)
                    \STATE $\mathcal{B} \leftarrow \mathcal{B} \cup \{\mathbf{S}, \mathbf{S}^\prime, \mathbf{p}_{\text{ref}}, r, done \}$
                \ELSE 
                    \STATE $\textcolor{red}{r_p}$ = Evaluation (\ref{eq:priv_cost}) with $\{ \mathcal{T}_{\text{opt}}, \mathcal{M} \}$
                    \STATE $\{\mathbf{S}, \mathbf{S}^\prime, \mathbf{p}_{\text{ref}}, \_, done \} \leftarrow$ $\mathcal{M}$.Step($\mathcal{T}_{\text{opt}}[0].\mathbf{u}$)
                    \STATE $\mathcal{B} \leftarrow \mathcal{B} \cup \{\mathbf{S}, \mathbf{S}^\prime, \mathbf{p}_{\text{ref}}, \textcolor{red}{r_p}, done \}$  \\  \COMMENT{Privileged reward}
                \ENDIF
                \IF{done}
                    \STATE $episode$ += 1
                    \STATE CurriculumUpdate($\mathcal{M}$, $episode$)
                \ENDIF
            \ENDFOR
            \STATE Training $\{ \theta, \phi \}$ with RL algorithm using $\mathcal{B}$, $\mathcal{B} \leftarrow \emptyset$
        \ENDWHILE
        \RETURN $\theta$ 
    \end{algorithmic}
\end{algorithm}

\section{Simulation Experiments}
\subsection{Experiment Setup}
\subsubsection{Simulation environment}
\label{sec:env}
Benefits from the agent-level feature representation in the proposed method, high-fidelity simulation environments are not required for network training. To ensure fairness in the comparative experiments, we use the same simulation environment as existing works \cite{zhou2022navigating, huajian2024sample}, with a brief description provided here: This 2D environment simulates a dynamic scene with heterogeneous obstacle constraints, wherein the ego-robot navigates through a corridor constrained by walls on both sides with a separation of 10 m. Along the way, a large rectangle obstacle (H, W $\in$ [1.0, 3.0] m) and several smaller static circular obstacles (R $\in$ [0.1, 0.4] m) are placed, while a crowd of pedestrians traverse freely through the obstacles. The sampling interval of the environment $\Delta t$ is set to 0.25 s, and the timeout threshold $T$ is set to 30 s, allowing a maximum of 120 interaction steps per episode.

\textbf{Crowd simulation}: All pedestrians are controlled by a centralized ORCA-based planner, which can avoid collisions among pedestrians and static obstacles. The pedestrians' initial and goal positions are roughly symmetrically placed with respect to the origin of the world coordinate. To simulate a continuous human flow, a new goal will be immediately generated once a pedestrian arrives at its goal position.

\textbf{Robot kinematic}: The kinematic model of the robot (\ref{eq:dyna}) is discretized using the explicit forward Euler method within the simulation environment. The maximum velocity $v_{\max}$ and acceleration $a_{\max}$ are set to 1.0 m/s and 1.0 m/s$^2$, respectively. We employ a hard-clipping policy to address violations of the kinematic constraints. In each training or validation episode, the initial position of the robot is fixed and its goal position is randomly sampled in $\mathcal{W}$. In test episodes, both the initial position and navigation goal of the robot are fixed.

\textbf{Curriculum setup}: The entire training process is divided into four phases. 1) The robot is given a random goal within the corridor with only one randomly placed static circular obstacle and a pedestrian; 2) Another two static circular obstacles and the static large rectangle obstacle are added, the robot's goal is limited on the line of $y = 4.0$; 3) Another two pedestrians are added; 4) The pedestrian number is increased to 5 to reflect more complex interactions. The training process continues for 50k episodes in total, with 4k, 8k, 8k, and 30k episodes allocated to each phase, respectively.
\subsubsection{Evaluation metrics}
We employ four key metrics to quantitatively evaluate the overall performance of the proposed method. Within environments characterized by dynamic interactions between humans and robots, safety is the foremost priority for autonomous navigation tasks. Accordingly, the primary objective of the planner is to minimize the percentage collision rate (CR). However, as task failures can also result from situations such as the robot getting stuck and timing out, we adopt its dual metric, success rate (SR), as the primary indicator to assess the planner’s ability to safely complete navigation tasks. Average navigation time (NT) is another key metric that evaluates navigation efficiency. Besides assessing navigation performance using these three metrics, we use the number of intrusions into humans’ social comfortable distance (DN) as a social metric. DN quantitatively measures the planner’s social compliance, which is the second most important criterion in crowd navigation. A high DN value indicates that the robot significantly disrupts pedestrian behavior, violating the fundamental principle that robots should serve rather than interfere with humans.
\subsubsection{Baseline methods}
We select eight existing open-source methods as benchmarks for performance comparison, namely A*+DWA, NH-ORCA, TEB, MPC-D-CBF, SG-D3QN, RL-RVO, HGAT-DRL, and SAGE. Among them, A*+DWA is the most classic solution for robot navigation, and its performance metrics provide an intuitive and quantifiable measure of the navigation task's difficulty. NH-ORCA represents model-based methods, extending ORCA by incorporating nonholonomic constraints of robots. Both TEB and MPC-D-CBF are optimization-based methods. The former is a well-established traditional approach, while the latter is opted to represent recent non-learning methods. 

Among the remaining four recent agent-based DRL methods, SG-D3QN, HGAT-DRL, and our previous work, SAGE, adopt graph-based feature extractors, similar to the proposed method. In contrast, RL-RVO utilizes a bidirectional gated recurrent unit (BiGRU) module for feature aggregation. From a pipeline perspective, SG-D3QN employs explicit online planning with a finite number of deduction steps, while the other learning-based methods generate motion commands end-to-end. The most notable distinction between the proposed method and these learning-based approaches is that our method outputs a local goal instead of direct robot control commands.
\subsubsection{Implementation details}
We implemented the front-end neural network based on the Stable-Baselines3 framework and constructed the GNN using the Deep Graph Library (DGL). The backend was developed in C++ and communicates with the Python-based front-end via ROS. We employed CasADi for modeling the optimization problem (\ref{eq:mpc}) and used Ipopt to solve it during training. During testing, we fixed the hyperplanes based on the reference trajectory and used Forces Pro to solve the regularized problem iteratively, ensuring the real-time performance of the planner. The stage number of MPC was set to $N$ = 10, with a timestep $\Delta t$ = 0.25~s, aligned with the sampling interval of the simulation environment. The proposed planner was end-to-end trained using the widely adopted Proximal Policy Optimization (PPO) algorithm with clipped gradients \cite{schulman2017proximal}, and all learnable parameters were updated using the Adam optimizer. For a fair comparison, we tested all methods with 500 random unseen test cases, and each of the five learning-based methods was trained five times with different random seeds. All training and experiments were conducted on an Ubuntu PC equipped with an Intel Core i9 CPU, 32 GB RAM, and an NVIDIA RTX 3080 GPU. A summary of all hyperparameters is provided in Table \ref{tab:hyperparams}. All unspecified hyperparameters retain their default values in Stable-Baselines3.

\begin{table}[htbp]
    \caption{The hyper-parameters of the training process\label{tab:hyperparams}}
    \centering
    \begin{tabular}{l l || l l}
        \toprule 
        \textbf{Parameter} & \textbf{Value} & \textbf{Parameter} & \textbf{Value} \\
        \midrule
        Initial learning rate & 2.5e-4  & $\kappa_1$ &  25.0 \\ 
        Rollout buffer size & 2048 & $\kappa_2$ & -25.0 \\
        Entropy coef. & 0.001 & $\kappa_3$ & 0.3 \\
        Traj. evaluation step $M$ & 4 & $\kappa_4$ & 1.0 \\
        Traj. discount factor $\lambda_p$ & 0.9 & $\kappa_5$ & -3.0 \\ 
        $\mu_s$ & 0.2~(m) & $\kappa_6$ & -1.5 \\
        $d_{\psi}$ & 5.0 & $\kappa_7$ & 50 \\
        $\xi_c$ & 3.0~(s) & $\kappa_c$ & $0.01$ \\
        \bottomrule
    \end{tabular}
\end{table}
\subsection{Quantitative Results}
\subsubsection{Comparison with baselines}
We summarize the statistical results of the comparative experiments in Table \ref{tab:baseline_compare}. Since non-learning-based methods are deterministic, they are tested only once, and no standard deviation is reported. As expected, the proposed method achieves the highest success rate of 99.4\% and the lowest collision rate of 0.5\%. More importantly, the proposed method also achieves the lowest DN, indicating its ability to effectively reduce disturbances to pedestrians and enhance social awareness during navigation. In terms of average navigation time, the proposed method is approximately 3.24 seconds slower than the optimal value among the baselines (13.33 seconds from RL-RVO), resulting in an efficiency loss of around 20\%. 

\begin{table}[htbp]
    \caption{Statistical comparison of different methods (mean/sd)\label{tab:baseline_compare}}
    \centering
    \resizebox{\columnwidth}{!}{
    \begin{tabular}{l l l l l}
        \toprule 
        \textbf{Method} & SR $\uparrow$ & CR $\downarrow$ & NT $\downarrow$ & DN $\downarrow$ \\
        \midrule
        A*+DWA \cite{fox1997dynamic} & 0.35 & 0.65 & \pmb{11.25} & 1741 \\
        NH-ORCA \cite{alonso2013optimal} & 0.76 & 0.18 & 17.61 & 499 \\
        TEB \cite{rosmann2017integrated} & 0.84 & 0.13 & 15.92 & 500 \\
        MPC-D-CBF \cite{jian2023dynamic} & 0.84 & 0.16 & 12.82 & 761 \\ 
        \midrule
        SG-D3QN \cite{zhou2022robot} & 0.68/0.11 & 0.22/0.04 & 18.60/0.52 & 216.00/32.79 \\
        RL-RVO \cite{han2022reinforcement} & 0.86/0.02 & 0.14/0.02 & \pmb{13.33/0.35} & 253.00/119.16 \\
        HGAT-DRL \cite{zhou2022navigating} & 0.92/0.02 & 0.08/0.02 & 13.50/0.28 & 243.40/64.55 \\
        SAGE \cite{huajian2024sample} & 0.98/0.01 & 0.02/0.01 & 14.14/0.14 & 110.80/32.51 \\
        \midrule
        Ours & \pmb{0.994/0.002} & \pmb{0.005/0.002} & 16.57/1.28 & \pmb{23.00/5.78} \\
        \bottomrule
    \end{tabular}}
\end{table}

It is reasonable to attribute the lower success rate of RL-RVO to its inability to accurately evaluate the potential collision risks posed by non-reciprocal pedestrians, leading to an overly aggressive behavioral policy. This assertion is reinforced by the performance of A*+DWA, which does not explicitly consider dynamic obstacles and achieves the shortest navigation time of 11.25 seconds. Similarly, we contend that the proposed method outperforms HGAT-DRL due to similar reasons. Compared to our prior end-to-end learning-based approach, SAGE, the proposed method achieves comparable or slightly superior performance. However, as emphasized, the optimization-based backend is capable of explicitly enforcing constraints, ensuring the planner's safety even in out-of-distribution corner cases.
\subsubsection{Ablation study}
\label{subsec:ablation}
To intuitively demonstrate the functionality of the components within the proposed method, targeted ablation experiments were conducted based on the primary contributions outlined in this paper. We summarize the statistical results of these experiments in Table \ref{tab:ablation}. It is evident from the results that the incremental action mask has the most significant impact on the final performance, as it directly determines whether the learning process can successfully converge. Without this module, the sampled local goals during the exploration of RL may be located inside structured obstacles, rendering the backend optimization problem infeasible. Moreover, this issue cannot be rectified by adding penalty terms in the reward function \cite{grams2023dynamic}.

The privileged learning also plays a crucial role by effectively improving the temporal consistency of the trajectory, thereby reducing trajectory oscillations between consecutive planning steps. In contrast, the contribution of spatio-temporal joint search to overall performance improvement is relatively minor. We attribute this observation to the fact that a well-trained network can effectively guide the robot away from densely interactive pedestrian regions, limiting the impact of this module to a few extreme scenarios.

To validate this proposition, we designed an additional set of ablation experiments. The entire frontend network was removed, and the A* algorithm was utilized to generate a reference trajectory, considering only structured obstacles. Subsequently, the same MPC-based backend was employed for trajectory refinement, and the local goal was selected from the reference trajectory based on the length of the MPC prediction horizon and the robot’s reference velocity. The experimental results, as summarized in Table \ref{tab:ablation}, clearly show that without the guidance of the frontend network, the backend must handle complex interactions more frequently. The S-T joint search module significantly enhances the navigation success rate by 8\% without sacrificing efficiency, demonstrating its effectiveness.

Furthermore, we investigated the impact of the discretization resolution of the robot’s neighborhood on overall performance, with the results presented in Table \ref{tab:point_num}. As the discretization resolution increases, the planner’s performance improves correspondingly and stabilizes after exceeding 81 discrete points. Based on this observation, we set the action space dimension of the policy network to $D = 81$ for all other experiments.

\begin{table}[htbp]
    \caption{Comparison of different discretization resolutions (mean)\label{tab:point_num}}
    \centering
    \begin{tabular}{l | l l l l}
        \toprule 
        \textbf{Action Dimension} & SR $\uparrow$ & CR $\downarrow$ & NT $\downarrow$ & DN $\downarrow$ \\
        \midrule
        $D = 25$ & 0.938 & 0.010 & 18.98 & 51 \\
        $D = 49$ & 0.958 & 0.034 & 18.09 & 117 \\
        \rowcolor{lightgray!50}
        $D = 81$ & \pmb{0.994} & 0.005 & \pmb{16.57} & 23 \\
        $D = 121$ & 0.992 & \pmb{0.000} & 17.53 & \pmb{11} \\ 
        $D = 169$ & 0.990 & 0.008 & 18.61 & 48 \\ 
        \bottomrule
    \end{tabular}
\end{table}

\begin{table*}[htbp]
    \caption{The statistical results of ablation experiments (Mean$\pm$SD)\label{tab:ablation}}
    \centering
    \begin{tabular}{c | c c c | l l l l}
        \toprule 
        \textbf{Method} & \textbf{Action Mask} & \textbf{S-T Joint Search} & \textbf{Privileged} & SR $\uparrow$ & CR $\downarrow$ & NT $\downarrow$ & DN $\downarrow$ \\
        \midrule
        \multirow{5}{*}{Ours hybrid} & $\checkmark$ & $\checkmark$ & $\checkmark$ & \pmb{0.994$\pm$0.002} & \pmb{0.005$\pm$0.002} & 16.57$\pm$1.28 & \pmb{23.00$\pm$5.78} \\
        & $\checkmark$ & & $\checkmark$ & 0.979$\pm$0.005 & 0.009$\pm$0.003 & 16.61$\pm$1.30 & 32.50$\pm$10.06 \\
        & $\checkmark$ & $\checkmark$ & & 0.966$\pm$0.005 & 0.023$\pm$0.012 & \pmb{14.70$\pm$0.62} & 129.75$\pm$10.76 \\
        & $\checkmark$ & & & 0.958$\pm$0.013 & 0.034$\pm$0.011 & 14.91$\pm$0.66 & 158.50$\pm$30.76 \\
        \rowcolor{lightgray!50}
        & & $\checkmark$ & $\checkmark$ & -- & -- & -- & -- \\
        \midrule
        \multirow{2}{*}{MPC only} & & $\checkmark$ & & 0.830 & 0.170 & \pmb{11.35} & 1169 \\
        & & & & 0.766 & 0.234 & 11.46 & 1108 \\ 
        \bottomrule
    \end{tabular}
\end{table*}
\subsubsection{Generalization capability}
\begin{figure}[htbp]
    \centering
    \subfloat{\includegraphics[width=0.75\columnwidth]{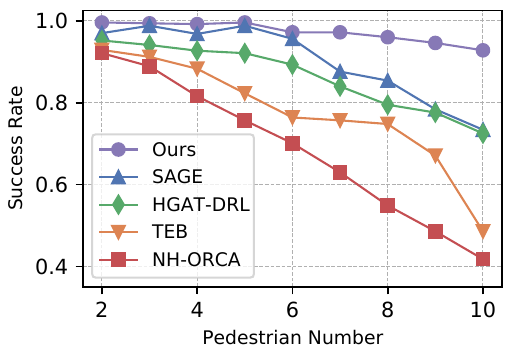}} \\
    \subfloat{\includegraphics[width=0.75\columnwidth]{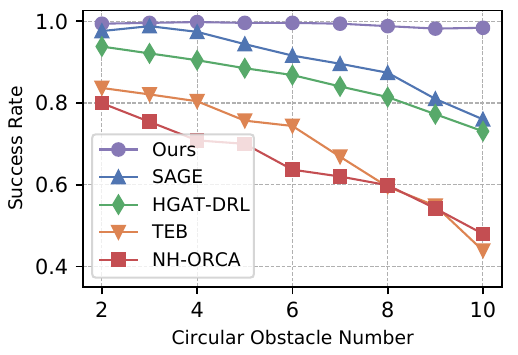}} \\
    \caption{Comparative experimental results of the proposed method against three baseline methods with variations in crowd size. The number of static circular obstacles is fixed to 3 when altering the number of pedestrians and the number of humans is fixed to 5 when the number of static circular obstacles is changed.
    \label{fig:generalization}}
\end{figure}

To better demonstrate the inherent adaptability of the graph-based representation to crowds of
different sizes, we conducted a comparative analysis involving the proposed method and four baseline methods: NH-ORCA, TEB, HGAT-DRL, and SAGE in scenarios featuring 2 to 10 pedestrians or static circular obstacles. The experiment results, quantitatively evaluated using the most representative metric SR, are depicted in Fig. \ref{fig:generalization}. It is evident from the data that the proposed method exhibits superior adaptability to variations in crowd size. Its performance advantage over the baselines expands as the scene complexity increases, whereas the performance of other methods degrades significantly with a growing number of agents. Furthermore, the success rate of two traditional methods decreases significantly faster than that of learning-based methods. This highlights the limitations of traditional approaches in handling highly interactive scenarios.

\subsection{Qualitative Evaluation}
\begin{figure*}[htb]
    \centering
    \includegraphics[width=0.95\textwidth]{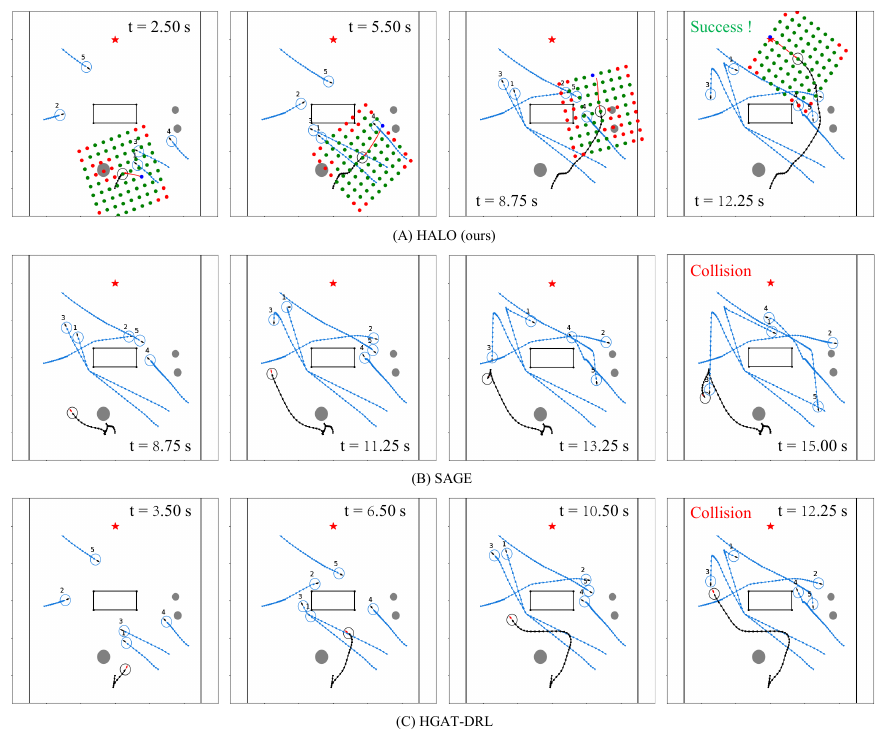}
    \caption{Qualitative comparison of the navigation process in representative simulation scenario. The ego-robot is depicted in black hollow circle with a red arrow indicates its heading direction and a red asterisk marks its goal. Pedestrians are depicted in blue hollow circles, static obstacles are depicted in gray solid circles and structured obstacles such as walls are plotted in black thin lines. Specifically for HALO, blue dots represent the local goals generated by the frontend network, red dots denote the invalid local goals masked out by the action mask, and red thin lines correspond to the locally optimal trajectories generated by the backend module.
    \label{fig:sim_visual}}
\end{figure*}

To intuitively demonstrate the characteristics of the proposed method, several representative simulation scenarios are selected to visualize the navigation processes of HALO and two baseline methods, SAGE and HGAT-DRL, as shown in Fig. \ref{fig:sim_visual}. A qualitative analysis of the illustrated navigation processes reveals that, while all three methods exhibit a certain level of scene understanding and can select less interactive routes based on pedestrian motion trends, only the proposed method is capable of correctly guiding the robot to bypass obstacles from the right and successfully reach the goal with a smooth and highly consistent trajectory in the complex scenarios depicted.

\section{Hardware Experiments}
\subsection{Hardware and Software Components}
The proposed method was implemented in a custom four-wheel skid-steer robot to demonstrate its real-world feasibility. To ensure consistency and minimize perception-related disturbances, LiDAR-based SLAM technology \cite{shan2020lio} was employed to construct an offline map containing only the static obstacles. During experiments, robot’s kinematic states were obtained through relocalization on the map at an update of 400~Hz. For pedestrians, we leveraged an external monitoring system based on recent work LV-DOT \cite{xu2025lv}, using a stereo camera (ZED2i) and a LiDAR (Livox Avia) to perform 3D dynamic object detection. This component served as a low-cost alternative to multi-camera BEV approaches. The perception system was deployed on a Jetson Orin AGX and communicated wirelessly with the robot via distributed ROS.

To enhance tracking accuracy, we employed a Kalman filter to track detected agents in a top-down 2D space, enabling high-frequency interpolation and delay compensation. Additionally, we incorporated an unsupervised re-identification (ReID) module \cite{Fu_2021_CVPR} to mitigate occlusion-related tracking inconsistency. For structured obstacles, we combined contour extraction with convex polygon decomposition \cite{liu2024optimization}, converting raw point-clouds into 2D polygon obstacles. As discussed earlier, to maintain real-time performance, we solved the optimization problem (\ref{eq:mpc}) using the Forces Pro solver at a replanning frequency of 20 Hz. Once the planned trajectory was obtained, we employed a 100 Hz cascaded PID controller to execute closed-loop trajectory tracking, utilizing high-frequency state feedback from the SLAM module.
\subsection{Comparative Results}
With a relatively high degree of repeatability, we designed a corridor-like experimental scenario consisting of one large polygonal obstacle, five static circular obstacles, and three pedestrians. The distance between the initial and target positions was set to 8 m, consistent with the simulation environment. The proposed method was evaluated against SAGE and HGAT-DRL, with each method tested 20 times. Statistical results are summarized in Table \ref{tab:exp_compare}. To better highlight the planner’s capabilities, the experimental scenarios are intentionally designed to be more challenging than typical daily-life environments. Furthermore, the participating pedestrians are instructed to avoid actively yielding to the robot whenever possible. It is important to note that, due to the inherent non-reproducibility of pedestrian behaviors, these experimental results, while providing an intuitive reference, should not be considered strictly rigorous.

\begin{table}[htbp]
    \caption{Experimental comparison of different methods (mean/sd)\label{tab:exp_compare}}
    \centering
    \begin{tabular}{l | l l l l}
        \toprule 
        \textbf{Method} & SR $\uparrow$ & CR $\downarrow$ & NT $\downarrow$ & DN $\downarrow$ \\
        \midrule
        HGAT-DRL \cite{zhou2022navigating} & 0.39 & 0.59 & 13.93/4.05 & 586 \\
        SAGE \cite{huajian2024sample} & 0.72	& 0.28 & \pmb{12.26/3.74} & 319 \\
        \rowcolor{lightgray!50}
        Ours & \pmb{0.84} & \pmb{0.16} & 15.57/2.65 & \pmb{277} \\
        \bottomrule
    \end{tabular}
\end{table}

The results of the real-world experiments were generally consistent with those observed in simulation. However, due to the increased spatial constraints of the experimental setup compared to the simulation environment, the DN values increased across all methods. We attribute the improvement of the proposed method over the baselines to the introduction of the optimization-based backend, which is capable of handling out-of-distribution scenarios, while the overall decline in performance across all methods, as compared to the simulation results, is likely due to limitations in the perception module.
\subsection{Quantitative Results}
\begin{figure*}[htbp]
    \centering
    \subfloat{\includegraphics[width=0.95\textwidth]{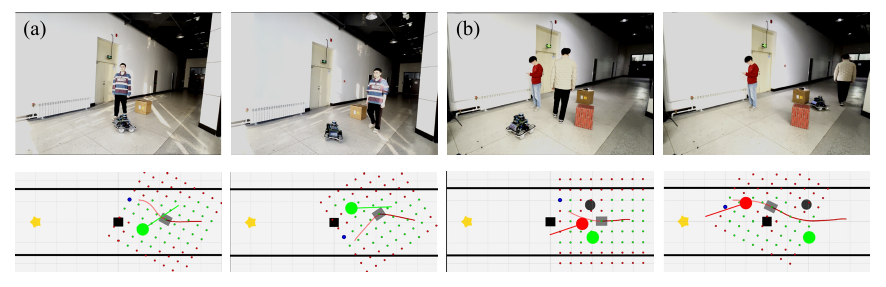}} \\
    \subfloat{\includegraphics[width=0.95\textwidth]{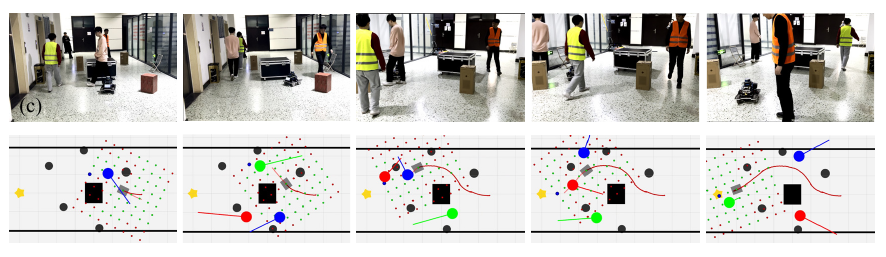}} \\
    \subfloat{\includegraphics[width=0.95\textwidth]{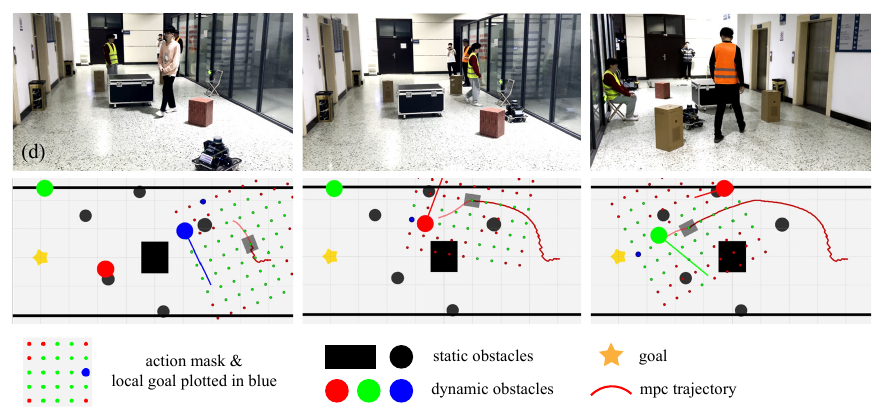}}
    \caption{Temporal visualizations of typical experimental scenarios, including synchronized perception and planning results. Thin lines with colors corresponding to the dynamic obstacles indicate their explicitly predicted future trajectories. More experimental results can be found in the video attachment.\label{fig:exp_vis}} 
\end{figure*}

To provide an intuitive illustration of the experimental results, several representative scenarios from the experiments were selected for visualization, as shown in Fig. \ref{fig:exp_vis}. First, a simple corridor scenario is used to illustrate the planner's anticipatory decision-making capability in Fig. \ref{fig:exp_vis}(a). Under identical initial and goal conditions, the planner is able to select different topological paths to bypass the obstacle from either side, based on the pedestrians’ motion trends, thereby reducing interactions. In Fig. \ref{fig:exp_vis}(b), a slow-moving pedestrian is walking in the same direction ahead of the robot, blocking its path. Additional obstacles further narrow the traversable area, limiting the robot’s options for detour. In this scenario, the robot is able to follow the slow pedestrian at a similar speed and resume normal speed to reach the goal once the path is cleared.

Fig. \ref{fig:exp_vis}(c) presents five sequential snapshots from a typical episode in the comparative experiments. Initially, with pedestrians present on both sides, the robot performs a backward motion to avoid interfering with the blue-marked pedestrian moving leftward and is guided to bypass the rectangular obstacle from the right side. Subsequently, when the right passage becomes occupied by a green-marked pedestrian, the robot proactively backs off to avoid collision, then handles interactions with two upcoming pedestrians by waiting. In the final snapshot, the robot performs another backward maneuver to avoid a pedestrian deliberately moving toward it and safely reaches the goal.

In Fig. \ref{fig:exp_vis}(d), three sequential snapshots depict another scenario. The planner first guides the robot to pass on the right to avoid interaction with the blue-marked pedestrian, based on the pedestrian’s motion trend. As the robot approaches a bench, its feasible path is blocked by a red-marked pedestrian with no possible detour. The robot slows down to wait until the pedestrian passes and sits down, then detours to the left to reach the goal. A similar deceleration is executed in the final snapshot, ensuring safe arrival at the destination.
\section{Conclusion and Discussion}
In this paper, we propose a novel local planning framework that integrates optimization and deep reinforcement learning for autonomous robot navigation in dynamic environments with structured constraints. By generating local goals, the S-T space is confined within the prediction horizon of MPC, enabling spatio-temporal joint search in open spaces to function similarly to that in road-like scenarios. More importantly, we design an incremental action masking mechanism, allowing the hybrid planner to be trained end-to-end in a low-fidelity simulation environment, thereby significantly reducing training costs. The proposed method achieves a 99\% success rate in a challenging simulation scenario, reaching SOTA performance. Real-world experiments on a physical robotic platform further demonstrate the deployment feasibility of the proposed method, as well as the sim-to-real generalization capability enabled by the GNN-based frontend.

We attribute the improvement of the proposed method over traditional optimization-based approaches to the implicit capability of the frontend network in estimating the future evolution of dynamic environments. Instead of relying on agile motion to maneuver through highly interactive crowds, the proposed planner proactively guides the robot away from potential dense interaction regions by local goal recommendation. On the other hand, we attribute its advantage over end-to-end learning-based methods to its robustness in out-of-distribution scenarios, particularly in bridging the sim-to-real gap. Under the situation where pedestrians can react to the robot, MPC ensures collision avoidance even if the robot fails to find a feasible trajectory. This further reinforces the navigation safety, as highlighted in the title. 

In real-world experiments, we also identified several limitations of the proposed method. First, the optimization-based backend does not explicitly account for the uncertainty of dynamic agents, which may lead the robot to adopt relatively extreme distances during unavoidable close-range interactions with pedestrians. This behavior can result in unexpected collisions under high perception noise. Second, we observed a certain degree of modal collapse, where the frontend network tends to favor one side of the corridor for routing while neglecting alternative optimal solutions. We attribute this phenomenon to the local optimum traps induced by the small rollout buffer in the PPO algorithm, which causes the learning process to reinforce a particular locally optimal modal while neglecting or forgetting other potential policies.

Therefore, our future work includes incorporating the impact of dynamic uncertainty into the backend optimization and mitigating modal collapse by leveraging off-policy algorithms such as Discrete SAC \cite{christodoulou2019soft}. Additionally, we plan to account for the robot's geometric shape in the planning framework, enabling the proposed method to be better suited for hardware platforms with larger aspect ratios, such as quadrupedal robots.
\section*{Acknowledgments}
The author(s) used ChatGPT for proofreading and manually reviewed the final manuscript.

\bibliographystyle{IEEEtran}
\bibliography{IEEEabrv,./ref}

\appendix
\section*{Additional Qualitative Results}
\begin{figure*}[htb]
    \centering
    \includegraphics[width=1.0\textwidth]{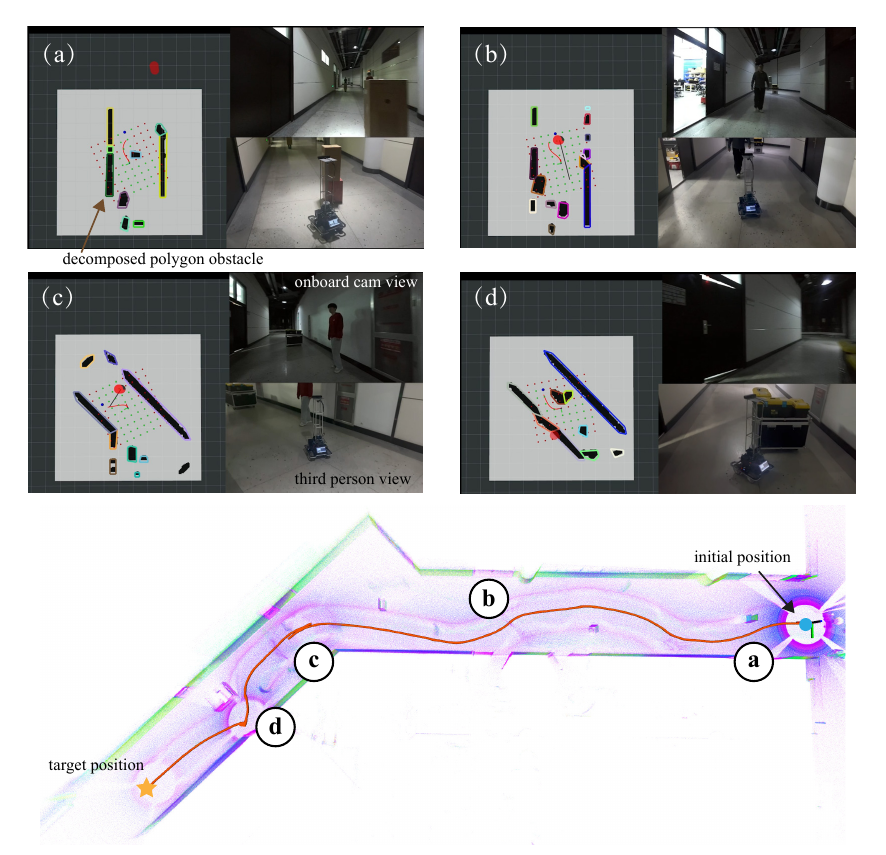}
    \caption{Visualized results of long-distance multi-point navigation experiments. The lettered labels in the figure indicate representative frames.\label{fig:exp_long}} 
\end{figure*}

To validate the feasibility of the proposed method in a realistic deployment setting using only onboard sensors and without any prior map, an additional qualitative long-range multi-goal navigation experiment is conducted. Unlike previous experiments that employed external perception modules, this experiment utilizes an onboard Livox Mid-360 LiDAR in combination with an RGB camera for real-time perception. Once pedestrians are detected, their corresponding point clouds are removed from the raw LiDAR data. The remaining point cloud is then projected onto a 2D plane to extract the occupancy map of static obstacles. Fig. \ref{fig:exp_long} presents keyframe snapshots of the navigation process along with the corresponding perception and planning visualizations. Additionally, a synchronized accumulated point cloud map is shown below to illustrate the spatial relationships between the snapshots.

\begin{figure*}[htb]
    \centering
    \includegraphics[width=0.95\textwidth]{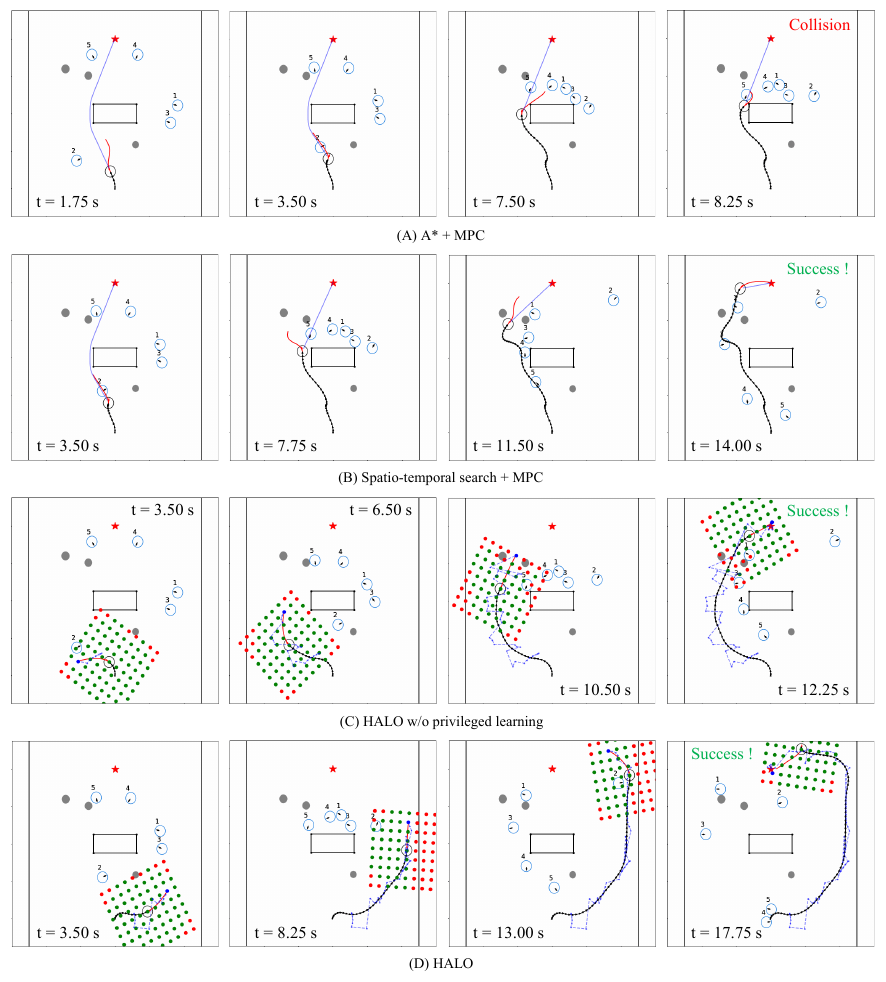}
    \caption{Qualitative comparison of the ablation models in representative simulation scenario. As an extension of the legend in Fig. \ref{fig:sim_visual}, the blue solid lines represent reference paths generated by the A* algorithm, while the blue dashed lines connect the goal points generated at each planning step to highlight their variation.\label{fig:ablation_visual}} 
\end{figure*}

As a supplement to the ablation study, representative scenarios are visualized to illustrate the impact of each sub-module of the proposed method on the navigation results, as shown in Fig. \ref{fig:ablation_visual}. Specifically, Fig. \ref{fig:ablation_visual}(a) and (b) intuitively demonstrate the benefits of joint spatio-temporal search, which provides more reasonable initializations for the backend optimization and significantly enhances the obstacle avoidance performance of the planner under dense interactions. The frontend decision network of HALO exhibits the ability to guide the robot away from densely interactive regions, as evidenced by the outer detours shown in Fig. \ref{fig:ablation_visual}(c) and (d). This behavior is consistent with the observations discussed in Section \ref{subsec:ablation}. Furthermore, a comparison between sub-figures (c) and (d) reveals that privileged learning improves the temporal consistency of the local goal points.
\end{document}